\documentclass{article}

\usepackage{microtype}
\usepackage{graphicx}
\usepackage{subfigure}
\usepackage{booktabs} 

\usepackage{silence}
\usepackage{hyperref}
\usepackage{caption}

\usepackage[accepted]{icml2024}

\usepackage{amsmath}
\usepackage{amssymb}
\usepackage{mathtools}
\usepackage{amsthm}
\usepackage[ruled,vlined]{algorithm2e} 

\usepackage[capitalize,noabbrev]{cleveref}

\theoremstyle{plain}

\theoremstyle{definition}

\theoremstyle{remark}

\usepackage[textsize=tiny]{todonotes}

\icmltitlerunning{EVCL: Elastic Variational Continual Learning with Weight Consolidation}

\begin{document}

\twocolumn[
\icmltitle{EVCL: Elastic Variational Continual Learning with Weight Consolidation}




\begin{icmlauthorlist}
\icmlauthor{Hunar Batra}{oxford}
\icmlauthor{Ronald Clark}{oxford}
\end{icmlauthorlist}

\icmlaffiliation{oxford}{Department of Computer Science, University of Oxford}

\icmlcorrespondingauthor{Hunar Batra}{hunar.batra@cs.ox.ac.uk}

\icmlkeywords{Bayesian Deep Learning, Continual Learning, Bayesian Neural Networks, Catastrophic Forgetting, ICML}

\vskip 0.3in
]



\printAffiliationsAndNotice{ }  

\begin{abstract}
Continual learning aims to allow models to learn new tasks without forgetting what has been learned before. This work introduces Elastic Variational Continual Learning with Weight Consolidation (EVCL), a novel hybrid model that integrates the variational posterior approximation mechanism of Variational Continual Learning (VCL) with the regularization-based parameter-protection strategy of Elastic Weight Consolidation (EWC). By combining the strengths of both methods, EVCL effectively mitigates catastrophic forgetting and enables better capture of dependencies between model parameters and task-specific data. Evaluated on five discriminative tasks, EVCL consistently outperforms existing baselines in both domain-incremental and task-incremental learning scenarios for deep discriminative models.
\end{abstract}

\section{Introduction}
Continual Learning focuses on training models on non-stationary data from sequential tasks, where tasks change over time or new tasks can emerge \cite{schlimmer1986case, sutton1993online, ring1997child}. The challenge is to balance between adapting to new data (plasticity) and retaining existing knowledge (stability). Given a sequence of tasks $D \in \{D_1, \ldots, D_T\}$, where each task $D_t = \{(x_i^t, y_i^t)\}_{i=1}^{N_t}$; $x_i^t \in \mathcal{X}$; $y_i^t \in \mathcal{Y}$, Continual Learning optimises a model $f_\theta: \mathcal{X} \rightarrow \mathcal{Y}$ parameterized by $\theta$, and aims to infer a distribution over predictive functions that performs well on the set of all tasks $D$, with access to only a single task at a time. However, when trained sequentially, neural networks (NNs) often suffer from catastrophic forgetting, where knowledge of previously learned tasks is abruptly lost as the model adapts to new tasks \cite{mccloskey1989catastrophic,ratcliff1990connectionist, goodfellow2013empirical, nguyen2019toward}.
  
Various approaches have been proposed to mitigate catastrophic forgetting, including inference based approaches \cite{nguyen2017variational, swaroop2019improving, kirkpatrick2017overcoming}, regularisation-based methods \cite{kirkpatrick2017overcoming, zenke2017continual, pan2020continual}, replay-based memory contextualisation \cite{rolnick2019experience, shin2017continual}, and architectural methods \cite{lee2017overcoming, rusu2016progressive, masse2018alleviating, loo2020combining}. Amongst the probabilistic inference methods, Variational Continual Learning (VCL) \cite{nguyen2017variational} and Elastic Weight Consolidation (EWC) \cite{kirkpatrick2017overcoming} are the two main approaches. VCL employs variational inference to approximate the posterior $p(\theta|D_t)$ based on a prior distribution $p(\theta)$. While, EWC penalises changes to parameters important for previous tasks by using the Fisher information matrix $F_t$ as a proxy for the posterior distribution $p(\theta|D_t)$, constraining the model to stay close to the optimal parameters $\theta_t^*$ for prior tasks.

Despite their individual strengths, both VCL and EWC have limitations in effectively mitigating catastrophic forgetting, with VCL suffering from accumulated errors due to approximate posterior alignment \cite{nguyen2017variational} and EWC's laplace approximation local estimate leading to underestimation of the importance of certain parameters \cite{huszar2018note, kirkpatrick2017overcoming}. This work introduces a novel integration of the variational inference mechanism of VCL with the EWC regularisation. Our proposed Variational Continual Learning with Weight Consolidation (EVCL) hybrid model combines the variational posterior approximation capabilities of VCL with the parameter-protection strategy of EWC, making it capture complex dependencies between model parameters and task-specific data better. Moreover, incorporating EWC's regularisation into VCL allows for direct optimization of the model parameters for each task by penalising changes to important parameters, as identified by the Fisher information matrix, mitigating the need for additional memory-intensive coresets. This work finds that EVCL effectively outperforms existing baselines on a variety of continual learning tasks for deep discriminative models. We release our code at \href{https://github.com/hunarbatra/elastic-variational-continual-learning}{https://github.com/hunarbatra/elastic-variational-continual-learning}. 



The contributions of this work are as follows:
\vspace{-1em}
\begin{itemize}
\setlength\itemsep{0.2em}
\item We propose Elastic Variational Continual Learning (EVCL), a novel hybrid method that integrates the strengths of Variational Continual Learning (VCL) and Elastic Weight Consolidation (EWC), combining the variational posterior approximation mechanism of VCL with the regularization-based parameter-protection strategy of EWC, effectively mitigating catastrophic forgetting.
\item EVCL outperforms existing baselines in both domain-incremental and task-incremental learning scenarios, achieving higher average test accuracies on a variety of continual learning tasks for deep discriminative models. 
\item EVCL demonstrates significantly less degradation in average accuracy as the number of tasks increases compared to other methods, underscoring its superior performance and stability in managing catastrophic forgetting. 
\end{itemize}

\section{Related Work}
A variety of approaches have been proposed to mitigate catastrophic forgetting, including regularisation, memory-based, and model architectural strategies. Our work on integrating VCL and EWC falls within the context of inference-based regularisation methods and parameter adaptation techniques, and falls within the broader landscape of regularisation, inference, and hybrid methods.

\textbf{Regularisation Approaches.} EWC \cite{kirkpatrick2017overcoming} and Synaptic Intelligence (SI) \cite{zenke2017continual} are foundational regularisation approaches that preserve critical parameters by penalising changes to significant weights. However, EWC's reliance on the Laplace approximation can lead to underestimation of parameter importance \cite{huszar2018note}. This work addresses this limitation by integrating VCL with EWC, providing a more robust framework for continual learning.

\textbf{Inference and Bayesian Approaches.} VCL \cite{nguyen2017variational} tackles catastrophic forgetting by approximating the posterior distributions of model parameters across sequential tasks. Likelihood-tempered VCL \cite{zhang2018noisy, osawa2019practical} modifies the approach by down-weighting the KL-divergence, but still faces challenges with posterior alignment. Our combined EVCL model enhances VCL's adaptation ability and provides a  better posterior approximation.

\textbf{Replay and Rehearsal Methods.} Replay-based methods, such as Experience Replay (ER) \cite{rolnick2019experience} and Deep Generative Replay (DGR) \cite{shin2017continual}, complement the current training data with representative data from past observations or generated pseudo-data. VCL with coreset \cite{nguyen2017variational} extends VCL by incorporating a coreset memory to regularise the variational posterior. Our approach is more memory-efficient and scalable as it does not require storing past data or generating pseudo-data.

\textbf{Hybrid Model Approaches.} Recent works have explored combining different approaches to improve continual learning performance. The Progress and Compress model \cite{schwarz2018progress} uses concepts from Progressive Neural Networks while employing EWC to protect important weights. Generalized VCL (GVCL) \cite{loo2020generalized} integrates multi-task FiLM architecture and likelihood-tempered variational inference, bridging VCL and Online EWC. Our approach aims for a computationally efficient integration of VCL and EWC without significantly increasing model complexity or memory demands.

\section{Preliminaries}

\subsection{Variational Continual Learning (VCL)}
VCL \cite{nguyen2017variational} is a Bayesian approach to continual learning that approximates the posterior distribution of the model parameters given the current task data and the accumulated approximate posteriors from previous tasks as the prior. The objective function of VCL for task $t$ is derived from the variational lower bound (ELBO) of the log-likelihood of the data (Appendix \ref{lem:elbo} Lemma 1), where we maximise the variational lower bound (i.e negative online variational free energy) to the online marginal likelihood, or equivalently minimising Kullback-Leibler divergence $\mathrm{KL}(\cdot \| \cdot)$:
\begin{multline}
\mathcal{L}_{\mathrm{VCL}}^t\left(q_t(\boldsymbol{\theta})\right) = \sum_{n=1}^{N_t} \mathbb{E}_{\boldsymbol{\theta} \sim q_t(\boldsymbol{\theta})}\left[\log p\left(y_t^{(n)} \mid \boldsymbol{\theta}, \mathbf{x}_t^{(n)}\right)\right] \\
- \mathrm{KL}\left(q_t(\boldsymbol{\theta}) \| q_{t-1}(\boldsymbol{\theta})\right)
\end{multline}
where $q_t(\boldsymbol{\theta})$ is the variational approximation of the posterior distribution at task $t$, $N_t$ is the number of data points in task $t$, $y_t^{(n)}$ and $\mathbf{x}_t^{(n)}$ are the target and input for the $n$-th data point in task $t$.

\subsection{Elastic Weight Consolidation (EWC)}
EWC \cite{kirkpatrick2017overcoming} is a regularization-based approach that preserves important parameters for previous tasks while allowing the model to adapt to new tasks. The EWC loss function is given by $\mathcal{L}_{\mathrm{EWC}}(\boldsymbol{\theta}) = \sum_i \frac{\lambda}{2} F_i^{t-1} \left(\theta_i - \theta_{t-1, i}^*\right)^2$, where $\lambda$ is a hyperparameter controlling the strength of the EWC regularization, $F_i^{t-1}$ is the Fisher information matrix computed from the previous task, $\theta_i$ are the current model parameters, and $\theta_{t-1, i}^*$ are the optimal model parameters for the previous task. The Fisher Information Matrix (FIM) $F_{t-1,i} = \mathbb{E}_{p(\mathcal{D}_{t-1} | \boldsymbol{\theta})} \left[ \left( \frac{\partial \log p(\mathcal{D}_{t-1} | \boldsymbol{\theta})}{\partial \theta_i} \right)^2 \right]$ captures the importance of each parameter $\theta_i$ for the previous task $t-1$. 

\section{Elastic Variational Continual Learning with Weight Consolidation (EVCL)}
Variational Continual Learning (VCL) approximates the posterior distribution of model parameters, capturing uncertainty and facilitating knowledge transfer across tasks. However, VCL suffers from catastrophic forgetting due to the divergence between the approximate and true posteriors, leading to accumulated errors. Additionally, VCL's reliance on coresets and additional episodic memory can limit its scalability and flexibility \cite{nguyen2017variational}.

To address these limitations, we propose Elastic Variational Continual Learning with Weight Consolidation (EVCL), which integrates Elastic Weight Consolidation (EWC) \cite{kirkpatrick2017overcoming} into the VCL framework. EWC is a regularisation-based approach that penalises changes to important parameters for previous tasks, as determined by the Fisher Information Matrix (FIM). By incorporating EWC's regularisation term into the variational objective, EVCL balances the performance on the current task with the retention of knowledge from previous tasks.

EVCL leverages the strengths of both techniques, using VCL to approximate the posterior distribution and EWC to identify and preserve important parameters. The parameter space is shared among tasks with controlled overlaps, allowing for a form of parameter space decomposition that shields critical parameters from large updates. This ensures that EVCL does not diverge significantly from the parameter configurations crucial for previously learned tasks, thereby reducing the compounding of approximation errors across tasks.

The Fisher Information Matrix with diagonal approximation in EWC captures the importance of each parameter for the previous tasks, guiding the variational posterior to retain crucial knowledge. By leveraging historical parameter importance, EVCL minimises the need for episodic memory. The regularisation provided by the Fisher Information also guides the variational approximation to remain closer to the true posterior, enhancing the overall fidelity of the model across multiple tasks. 

EVCL combines the variational approximation methods of VCL with the stability-enhancing regularisation strategies of EWC, addressing the drawbacks associated with each method when used independently. In the following section we describe how we integrate VCL and EWC into a unified objective function and then empirically demonstrating its effectiveness in mitigating catastrophic forgetting across various continual learning benchmarks for deep discriminative models.

\subsection{Approach}
This section presents our approach EVCL, combining VCL and EWC to mitigate catastrophic forgetting in continual learning. 

For EVCL, to integrate EWC into the variational framework of VCL, we add the EWC penalty term to the VCL loss function. The EWC penalty term is computed as the sum of the squared differences between the models current variational parameters and those of the previous task, weighted by the Fisher Information Matrix $F_i^{t-1}$ and the regularization strength $\lambda$. 

By adding the EWC penalty term to the VCL loss, we obtain the combined EVCL loss function:
\begin{multline}
\mathcal{L}_{\mathrm{EVCL}}^t\left(q_t(\boldsymbol{\theta})\right) = \mathcal{L}_{\mathrm{VCL}}^t\left(q_t(\boldsymbol{\theta})\right) + \\ \sum_i \frac{\lambda}{2} F_i^{t-1} \left[\left(\mu_{t,i} - \mu_{t-1,i}\right)^2 +  
\left(\sigma_{t,i}^2 - \sigma_{t-1,i}^2\right)^2\right]
\end{multline}


where $\mathcal{L}_{\mathrm{VCL}}^t\left(q_t(\boldsymbol{\theta})\right)$ is computed using a Gaussian mean-field approximate posterior $q_t(\theta) = \prod_{d=1}^D\mathcal{N}(\theta_{t,d};\mu_{t,d},\sigma_{t,d}^2)$ to allow analytical computation of the KL divergence and expectations. Here, $\mu_{t,i}$ and $\sigma_{t,i}^2$ represents the mean and variance of the variational posterior for parameter $\theta_i$ at task $t$ respectively; and $\mu_{t-1,i}$ and $\sigma_{t-1,i}^2$ represents the mean and variance of the variational posterior for parameter $\theta_i$ at the previous task $(t-1)$ respectively.

Expected log-likelihood is approximated using simple Monte Carlo and local reparameterisation \cite{kingma2015variational, salimans2013fixed, kingma2014stochastic}. At the first time step, $q_0(\theta)$ prior distribution is chosen as a multivariate Gaussian distribution \cite{graves2011practical, blundell2015weight, nguyen2017variational}. The integration of the Fisher Information Matrix in EVCL employs a diagonal approximation to efficiently capture parameter importance without significantly increasing computational complexity.

The combined loss function encourages the model to find a variational approximation that not only fits the current task well but also integrates information from previous tasks through the VCL term, while the EWC term penalizes significant deviations in parameters crucial for previous tasks, ensuring stability and retaining knowledge. 

The hyperparameter $\lambda$, which we set to 100, controls the strength of the EWC regularization for EVCL, allowing for a trade-off between plasticity and stability. Notably, there is no explicit weighting between the VCL and EWC components in the combined EVCL loss; instead, the balance is controlled implicitly through the variational framework and the regularization strength parameter $\lambda$ of EWC.

By incorporating the EWC penalty term into the variational framework of VCL, we effectively regularize the variational approximation to preserve the important parameters for the previous tasks while allowing the model to adapt to new tasks, mitigating catastrophic forgetting in continual learning scenarios. 

An overview of the full algorithm can be found in Appendix \ref{algo}.

\section{Experiments}
We apply the proposed EVCL framework to discriminative models, specifically fully-connected neural network classifiers, and evaluate its performance on five tasks: PermutedMNIST, SplitMNIST, SplitNotMNIST, SplitFashionMNIST, and SplitCIFAR-10. 

For PermutedMNIST, we test the domain incremental context of continual learning using a single-head MLP. For the other tasks, we employ a multi-head MLP with separate output heads but shared model parameters to test the incremental continual learning contexts. EVCL is compared against multiple baselines including VCL, VCL with Random Coreset, VCL with K-center coreset, EWC, and standalone Coreset models. 

The EWC penalty used in EVCL and standalone EWC model, is set to $\lambda$ = 100. The Fisher Information Matrix is estimated using 5000 samples from the data, and the coreset size is set to 200. We train each of these models for 100 epochs with a batch size of 256. During evaluation, after training on each task, we test on all tasks seen so far and compute the average test accuracies across these tasks, with results averaged over 3 runs.

For EVCL, during training, the loss function is minimized with respect to the variational parameters of $q_t(\boldsymbol{\theta})$ using the Adam optimizer \cite{kingma2014adam} with a learning rate of 1e-3. The gradients of the EWC term used in EVCL are computed using the reparameterization trick \cite{kingma2015variational, kingma2014stochastic}, enabling end-to-end training of the model. 

\subsection{Task setup and Results}
\textbf{PermutedMNIST}: This is a domain-incremental learning task consisting of labeled MNIST images whose pixels have undergone a fixed random permutation \cite{goodfellow2013empirical, kirkpatrick2017overcoming, zenke2017continual}. This simple setup tests the robustness of models to domain shifts. For this task, we use a single-head two-layer MLP with 100 hidden units per layer and ReLU activations. The average test accuracy after training on 5 tasks shows that EVCL achieves 93.5\%, significantly outperforming VCL at 91.5\%, VCL with Random-Coreset at 91.68\%, VCL with K-Center Coreset at 92\%, and EWC at 65\%.
\begin{figure}[ht!]
    \centering
\includegraphics[scale=0.385]{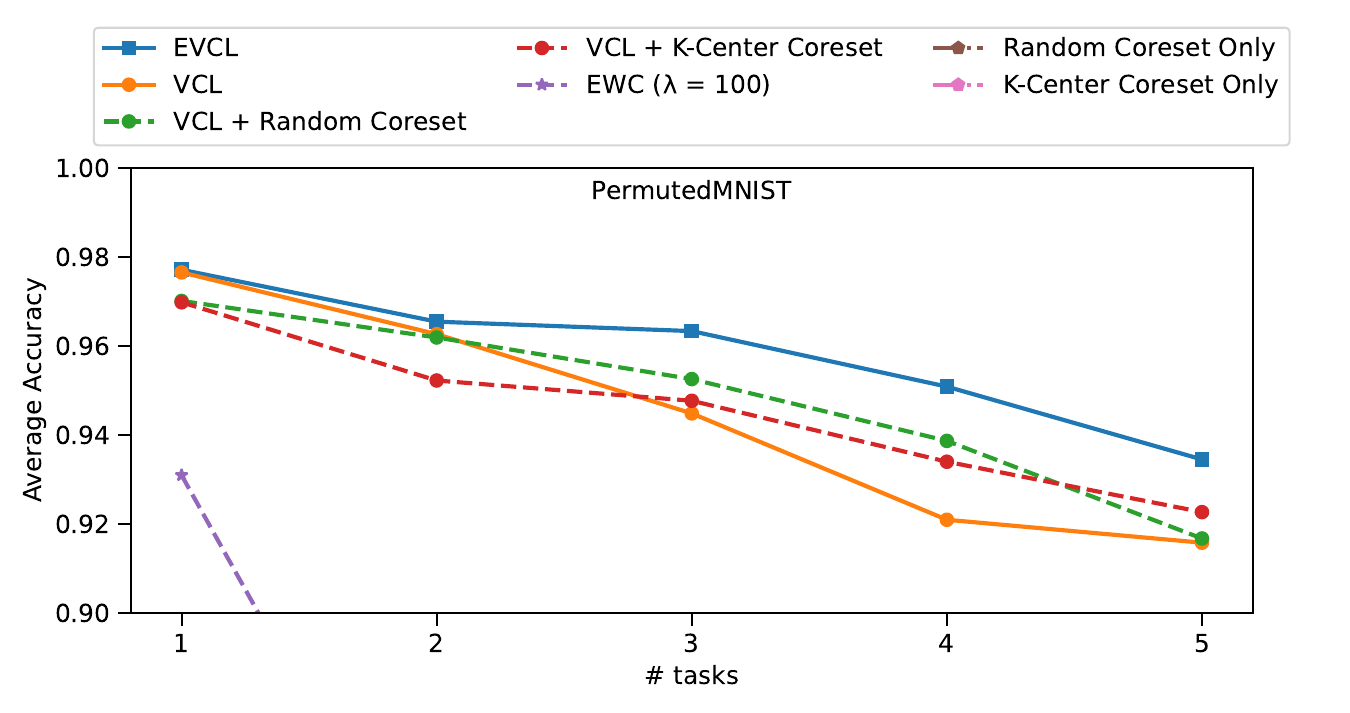}
\captionsetup{skip=1pt} 
\caption{Test set average accuracy over PermutedMNIST for EVCL and baseline models.}
\label{fig:permutedmnist}
\end{figure}

\textbf{SplitMNIST}: This task comprises sequential binary classification tasks derived from MNIST digits, specifically 0/1, 2/3, 4/5, 6/7, and 8/9 \cite{zenke2017continual}. A multi-head MLP network with 256 hidden units per layer is employed for this task. The average test accuracy after training on all 5 tasks in the SplitMNIST experiment shows that EVCL leads with 98.4\%, compared to VCL at 94\%, VCL with Random Coreset at 96\%, VCL with K-Center Coreset at 94.4\%, and EWC at 88\%.
\begin{figure}[ht!]
    \centering
\includegraphics[scale=0.385]{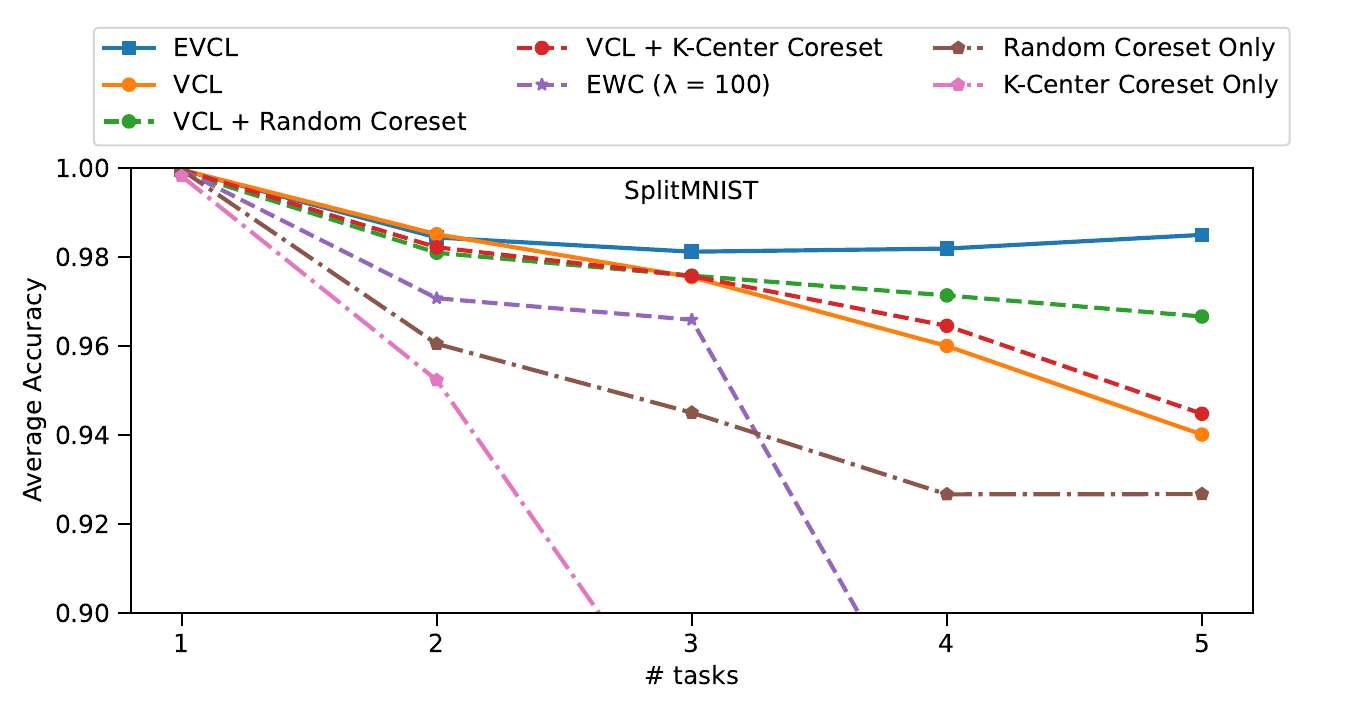}
\captionsetup{skip=1pt} 
\caption{Test set average accuracy over SplitMNIST for EVCL and baseline models.}
\label{fig:splitmnist}
\end{figure}

\textbf{SplitNotMNIST}: This experiment used by \cite{nguyen2017variational} challenges models with character classification from A to J across various fonts, split over 5 binary classification tasks: A/F, B/G, C/H, D/I, and E/J. A deeper multi-head network with four layers of 150 hidden units per layer and shared parameters is utilized. The average test accuracy after training on all tasks shows that EVCL records 91.7\%, outstripping VCL at 89.7\%, VCL with Random Coreset at 86\%, VCL with K-Center Coreset at 82.7\%, and EWC at 62.9\%. 
\begin{figure}[ht!]
    \centering
\includegraphics[scale=0.385]{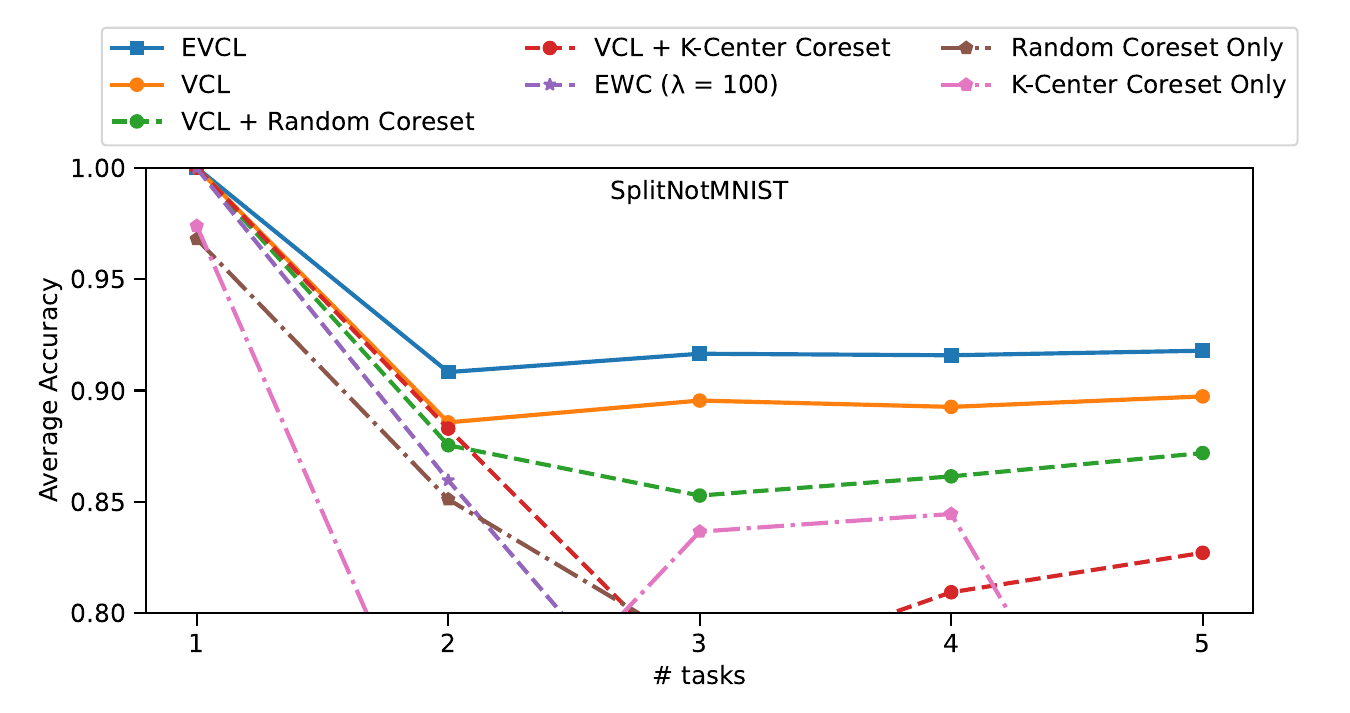}
\captionsetup{skip=1pt} 
\caption{Test set average accuracy over SplitNotMNIST for EVCL and baseline models.}
\label{fig:nomnist}
\end{figure}

\textbf{SplitFashionMNIST}: This experiment involves classifying fashion items into five categories: top/trouser, pullover/dress, coat/sandal, shirt/sneaker, and bag/ankle boots \cite{xiao2017fashion}. The architecture mirrors that of SplitNotMNIST, using a deeper multi-head network with four layers of 150 hidden units and shared parameters. The average test accuracy after training on all tasks shows that EVCL attains 96.2\%, exceeding VCL at 90\%, VCL with Random Coreset at 86\%, VCL with K-Center Coreset at 86.3\%, and EWC at 74\%.
\begin{figure}[ht!]
    \centering
\includegraphics[scale=0.385]{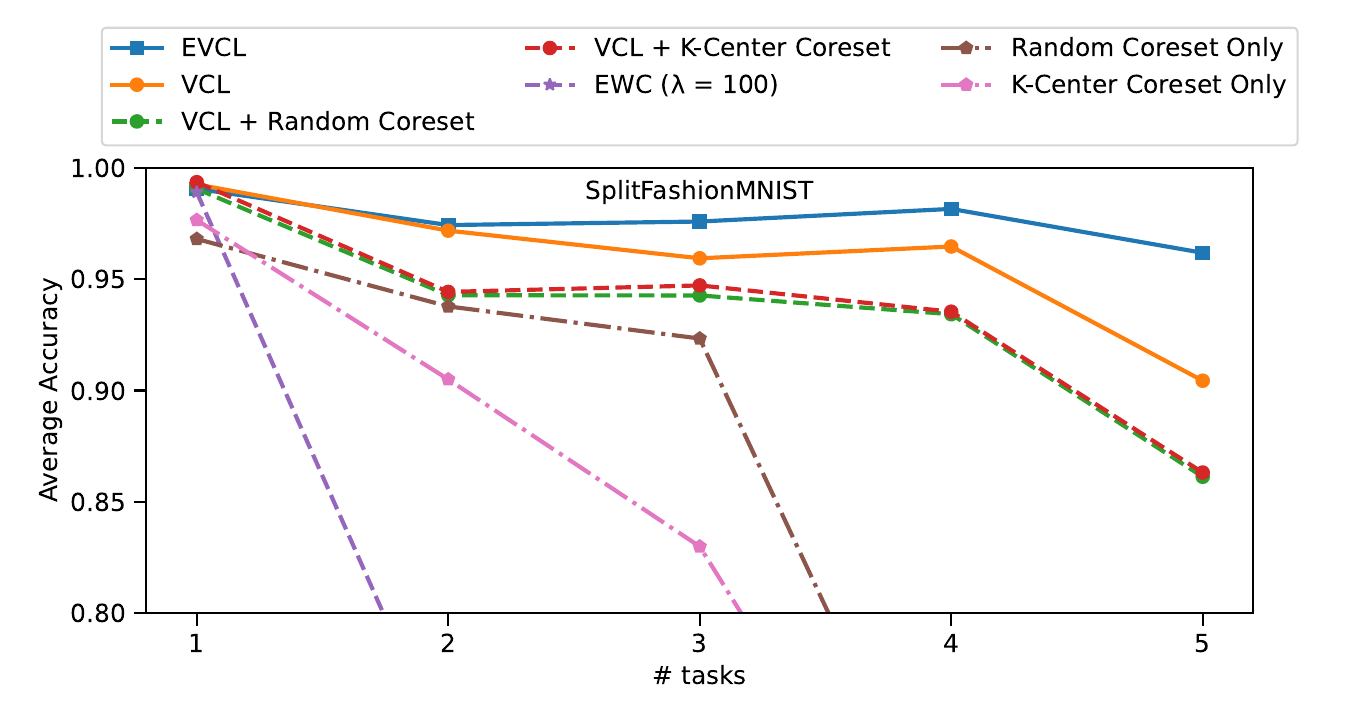}
\captionsetup{skip=1pt} 
\caption{Test set average accuracy over SplitFashionMNIST for EVCL and baseline models.}
\label{fig:fashionmnist}
\end{figure}

\textbf{SplitCIFAR-10}: This experiment tests classification across complex and diverse images such as airplanes, automobiles, birds, and cats, split over five binary classification tasks: airplane/automobile, bird/cat, deer/dog, frog/horse, and ship/truck \cite{krizhevsky2009learning}. The average test accuracy after training on all tasks shows that EVCL achieves 74\%, surpassing VCL at 72\%, VCL with Random Coreset at 71.5\%, VCL with K-Center Coreset at 67\%, and EWC at 59\%.
\begin{figure}[ht!]
    \centering
\includegraphics[scale=0.385]{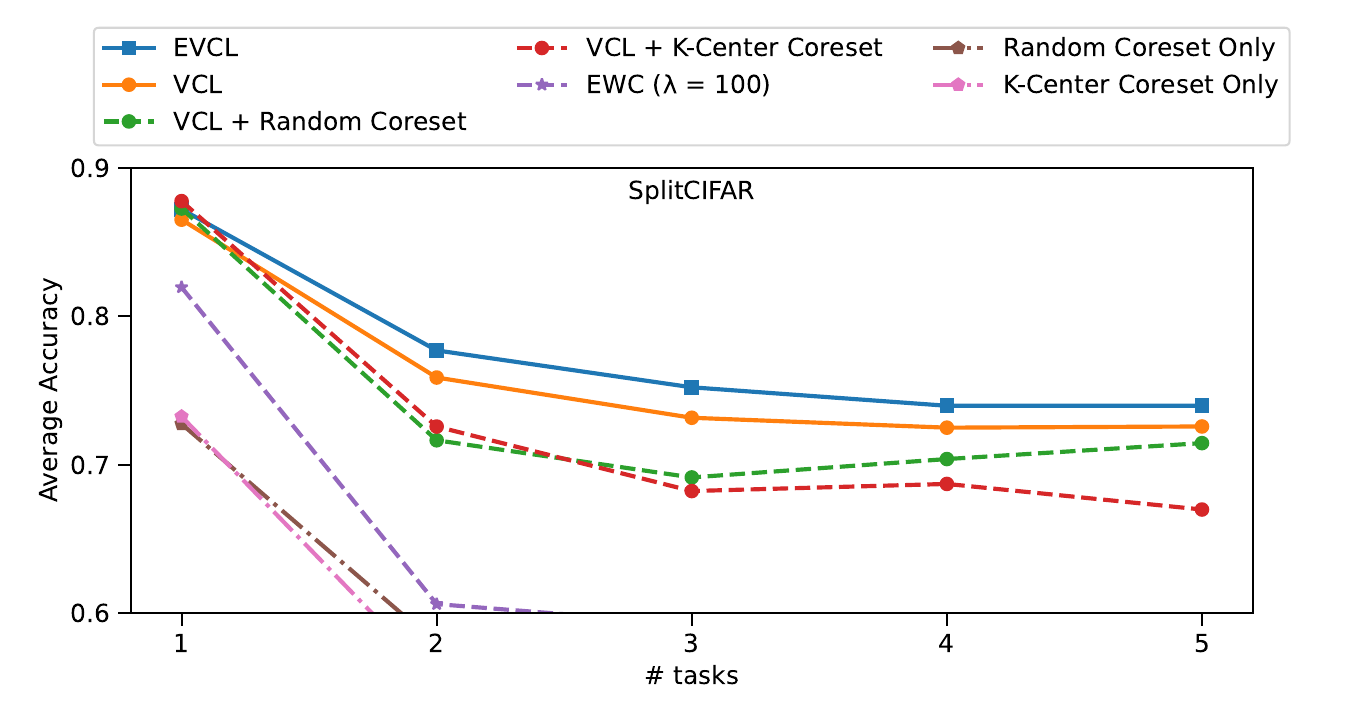}
\captionsetup{skip=1pt} 
\caption{Test set average accuracy over SplitCIFAR-10 for EVCL and baseline models.}
\label{fig:splitcifar}
\end{figure}

Across all tasks, EVCL consistently outperforms traditional VCL, VCL augmented with random and k-center coresets, EWC, and coreset-only approaches, demonstrating its effectiveness in managing catastrophic forgetting in complex continual learning scenarios. This improvement underscores the benefits of integrating EWC within VCL, especially in multi-task settings where model stability and adaptability are crucial. While all methods exhibit some degradation in accuracy as the number of tasks increases, EVCL shows significantly less degradation compared to the other methods, highlighting its robustness and superior performance.

\section{Conclusion}


Our proposed Elastic Variational Continual Learning (EVCL) model integrates Elastic Weight Consolidation (EWC) into the Variational Continual Learning (VCL) framework to mitigate catastrophic forgetting. EVCL consistently outperforms existing baselines for discriminative tasks by effectively balancing plasticity and stability. Additionally, EVCL exhibits significantly less degradation in performance as the number of tasks increases compared to other methods.

The model's performance could be further enhanced by incorporating natural gradient methods, such as Kronecker-factored Approximate Curvature (K-FAC) and Online Natural Gradient Descent \cite{martens2015optimizing, ollivier2018online}, to better approximate the Fisher Information Matrix (FIM) beyond the diagonal approximation, thus capturing the curvature of the parameter space more accurately and addressing EWC's limitations \cite{zhang2018noisy}.

Future work could explore the application of EVCL to generative models and reinforcement learning, as well as extending the framework to handle complex task structures and model architectures \cite{sankararaman2022bayesformer}. Additionally, incorporating experimental replay mechanisms \cite{rolnick2019experience, shin2017continual, nguyen2017variational}, sparse coding techniques \cite{sarfraz2023sparse}, and parameter efficient fine-tuning approaches \cite{hu2021lora, sharma2023truth} could potentially lead to more conclusive results in the field while improving the scalability and robustness of the approach.




\bibliography{example_paper}
\bibliographystyle{icml2024}

\newpage
\appendix
\onecolumn
\section*{Appendix}

\section{Elastic Variational Continual Learning with Weight Consolidation (EVCL) Algorithm}\label{algo}
\begin{algorithm}[H]
\caption{Elastic Variational Continual Learning with Weight Consolidation (EVCL)}
\SetAlgoLined
\KwIn{Dataset $\mathcal{D} = \{\mathcal{D}_1, \ldots, \mathcal{D}_T\}$, learning rate $\alpha$, EWC strength $\lambda$}
\KwOut{Variational parameters $\phi_t$ for each task $t$}
Initialize variational parameters $\phi_0$; Initialize prior $p(\boldsymbol{\theta} | \mathcal{D}_{0}) \leftarrow q_{\phi_0}(\boldsymbol{\theta})$\\
\For{$t = 1, \ldots, T$}
{
    Initialize $\phi_t \leftarrow \phi_{t-1}$\\
    \For{each batch $\mathcal{B} \subset \mathcal{D}_t$}
    {
        Compute VCL loss: 
        $\mathcal{L}_{\mathrm{VCL}}^t = \frac{1}{|\mathcal{B}|} \sum_{n=1}^{|\mathcal{B}|} \mathbb{E}_{q_{\phi_t}(\boldsymbol{\theta})}[\log p(y_n | \boldsymbol{\theta}, \mathbf{x}_n)] - \mathrm{KL}(q_{\phi_t}(\boldsymbol{\theta}) || p(\boldsymbol{\theta} | \mathcal{D}_{1:t-1}))$\\
        \eIf{$t > 1$}
        {
            Compute EWC loss: 
            $\mathcal{L}_{\mathrm{EWC}}^t = \sum_i \frac{\lambda}{2} F_i^{t-1} \left[ (\mu_{t,i} - \mu_{t-1,i})^2 + (\sigma_{t,i}^2 - \sigma_{t-1,i}^2)^2 \right]$
        }
        {
            $\mathcal{L}_{\mathrm{EWC}}^t = 0$\\
        }
        Compute total loss: 
        $\mathcal{L}_{\mathrm{EVCL}}^t = \mathcal{L}_{\mathrm{VCL}}^t + \mathcal{L}_{\mathrm{EWC}}^t$\\
        Update variational parameters: 
        $\phi_t \leftarrow \phi_t - \alpha \nabla_{\phi_t} \mathcal{L}_{\mathrm{EVCL}}^t$\\
    }
    Store model parameters mean and variance: 
    $\mu_{t-1,i} = \mathbb{E}_{q_{\phi_t}(\boldsymbol{\theta})}[\theta_i]$ and $\sigma_{t-1,i}^2 = \mathrm{Var}_{q_{\phi_t}(\boldsymbol{\theta})}[\theta_i]$ for all $i$\\
    Compute Fisher Information Matrix: 
    $F_i^t = \mathbb{E}_{p(\mathcal{D}_t | \boldsymbol{\theta})} \left[ \left( \frac{\partial \log p(\mathcal{D}_t | \boldsymbol{\theta})}{\partial \theta_i} \right)^2 \right]$ for all $i$\\
    Set prior for next task: 
    $p(\boldsymbol{\theta} | \mathcal{D}_{1:t}) \leftarrow q_{\phi_t}(\boldsymbol{\theta})$\\
}
\end{algorithm}

\section{Lemmas}
\textbf{Lemma 1 (Evidence Lower Bound):}\label{lem:elbo} Let $p(x | \theta)$ be the likelihood of the data $x$ given the parameters $\theta$, and let $q_\lambda(z)$ be a variational distribution parameterized by $\lambda$. The log marginal likelihood $\log p(x | \theta)$ can be decomposed as:
\begin{equation}
\log p(x | \theta) = \text{ELBO}_{\theta, \lambda} + \text{KL}(q_\lambda(z) || p(z | x, \theta)),
\end{equation}
where the Evidence Lower Bound (ELBO) is defined as:
$
\text{ELBO}_{\theta, \lambda} = \int q_\lambda(z) \log \frac{p(z, x | \theta)}{q_\lambda(z)} dz.
$

\textbf{Proof:}
Starting from the definition of the ELBO, we have:
\begin{align}
\log p(x | \theta) - \text{KL}(q_\lambda(z) || p(z | x, \theta)) &= \log p(x | \theta) \int q_\lambda(z) dz - \int q_\lambda(z) \log \frac{q_\lambda(z)}{p(z | x, \theta)} dz \\
&= \int q_\lambda(z) \log p(x | \theta) dz + \int q_\lambda(z) \log \frac{p(z | x, \theta)}{q_\lambda(z)} dz \\
&= \int q_\lambda(z) \log \frac{p(z, x | \theta)}{q_\lambda(z)} dz \\
&= \int q_\lambda(z) \log p(x, z | \theta) dz - \int q_\lambda(z) \log q_\lambda(z) dz \\
&= \text{ELBO}_{\theta, \lambda}.
\end{align}

Since $\text{KL}(\cdot || \cdot) \geq 0$, we have $\log p(x | \theta) \geq \text{ELBO}_{\theta, \lambda}$ for any $\theta$, $\lambda$, and $q_\lambda$. Hence, $\text{ELBO}_{\theta, \lambda}$ is called the Evidence Lower Bound, and
\begin{equation}
\arg\min_\lambda \text{KL}(q_\lambda(z) || p(z | x, \theta)) = \arg\max_\lambda \text{ELBO}_{\theta, \lambda}.
\end{equation}


\end{document}